\documentclass[runningheads]{llncs}
\usepackage{graphicx}
\usepackage{amsmath,amssymb} 
\usepackage{color}
\usepackage{pifont}
\usepackage{caption}
\usepackage{hyperref}
\newcommand{\cmark}{\ding{51}}
\newcommand{\xmark}{\ding{55}}

\newcommand{\ie}{\textit{i}.\textit{e}.}
\newcommand{\eg}{\textit{e}.\textit{g}.}

\begin{document}
\title{3D Pose Estimation for Fine-Grained\\Object Categories} 

\titlerunning{3D Pose Estimation for Fine-Grained Object Categories}
%
\author{Yaming Wang\inst{1} \and
Xiao Tan\inst{2} \and
Yi Yang\inst{2} \and
Xiao Liu\inst{2} \and
Errui Ding\inst{2} \and
Feng Zhou\inst{2} \and
Larry S. Davis\inst{1}}
%
\authorrunning{Y. Wang et al.}
%

\institute{University of Maryland, College Park, MD 20742, USA\\
\email{\{wym, lsd\}@umiacs.umd.edu} \and
Baidu, Inc.\\
\email{\{tanxiao01, yangyi05, liuxiao12, dingerrui, zhoufeng09\}@baidu.com}}
\maketitle              
\begin{abstract}
Existing object pose estimation datasets are related to generic object types and there is so far no dataset for fine-grained object categories. 
In this work, we introduce a new large dataset to benchmark pose estimation for fine-grained objects, thanks to the availability of both 2D and 3D fine-grained data recently. 
Specifically, we augment two popular fine-grained recognition datasets (StanfordCars and CompCars) by finding a fine-grained 3D CAD model for each sub-category and manually annotating each object in images with 3D pose. 
We show that, with enough training data, a full perspective model with continuous parameters can be estimated using 2D appearance information alone. 
We achieve this via a framework based on Faster/Mask R-CNN. 
This goes beyond previous works on category-level pose estimation, which only estimate discrete/continuous viewpoint angles or recover rotation matrices often with the help of key points. 
Furthermore, with fine-grained 3D models available, we incorporate a dense 3D representation named as {\em location field} into the CNN-based pose estimation framework to further improve the performance.
The new dataset is available at \url{www.umiacs.umd.edu/~wym/3dpose.html}
\end{abstract}

\section{Introduction}
In the past few years, the fast-pacing progress of generic image recognition on ImageNet~\cite{krizhevsky2012imagenet} has drawn increasing attention of research in classifying fine-grained object categories~\cite{krause2016unreasonable,van2017inaturalist}, \eg bird species~\cite{wah2011cub}, car makes and models~\cite{krause20133d}.
However, simply recognizing object labels is still far from solving many industrial problems where we need to have a deeper understanding of other attributes of the object~\cite{lim2013parsing}.
In this work, we study the problem of estimating 3D pose for fine-grained objects from monocular images.
We believe this will become an indispensable component in some broader tasks.
For example, to build a vision-based car damage assessment system, an important step is to estimate the exact pose of the car so that the damaged part can be well aligned for further detailed analysis.
\begin{figure}
\centering
\includegraphics[width=0.9\textwidth]{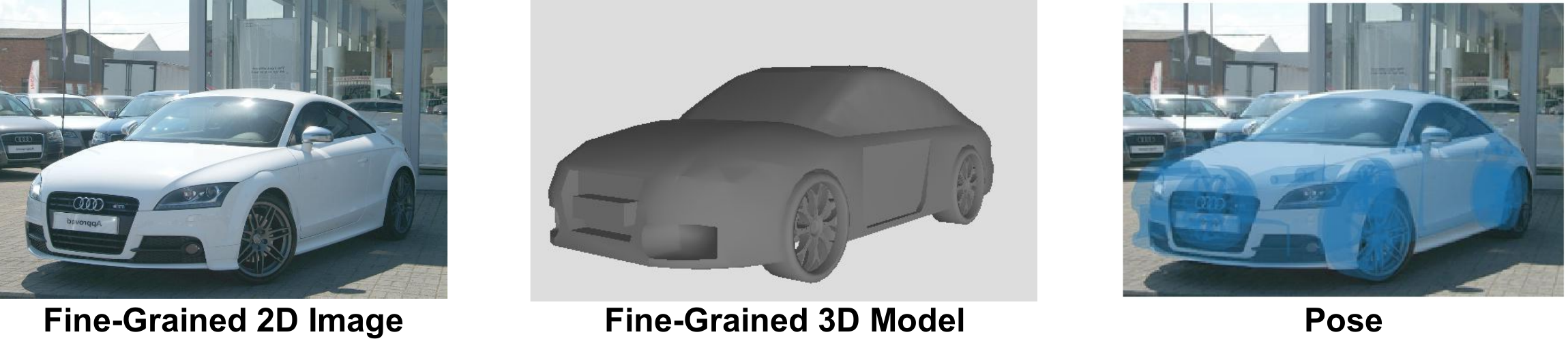}
\caption{\label{fig1}For an image from a fine-grained category (Left), we find its corresponding fine-grained 3D model (Middle) and annotate its pose (Right). The problem is to estimate the pose such that the projection of the 3D model align with the image as well as possible.}
\end{figure}

To address this task, collecting suitable data is of vital importance. However, large-scale as they are, recent category-level pose estimation datasets are typically designed for generic object types \cite{xiang2016objectnet3d,xiang2014beyond} and there is so far no large-scale pose dataset for fine-grained object categories.
Although datasets on generic object types could contain decent information for pose, they lack of fine-detailed matching of object shapes during annotation, since they usually use only a few universal 3D object models to match a group of objects with different shapes in one hyper class~\cite{xiang2014beyond}.
In this work, we introduce a new dataset that is able to benchmark pose estimation for fine-grained objects.
Specifically, we augment two existing fine-grained recognition datasets, StanfordCars~\cite{krause20133d} and CompCars~\cite{yang2015large}, with two types of useful 3D information: (\textit{i}) for each car in the image, we manually annotate the pose parameters for a full perspective projection; (\textit{ii}) we provide an accurate match of the computer aided design (CAD) model for each category. 
The resulting augmented dataset consists of more than 20,000 images for over 300 fine-grained categories.

To our best knowledge, this is the first work for fine-grained object pose estimation.
Given the built dataset with high-quality pose annotations, we show that the pose parameters can be predicted from a single 2D image with only appearance information.
Compared to most previous works~\cite{zhou20153d,tulsiani2015viewpoints,pavlakos20176}, our method does not require the intermediate prediction of 2D/3D key points. 
In addition, we assume a full perspective model, which is a more challenging setting than previous works of estimating discrete/continuous viewpoint angles (azimuth)~\cite{ghodrati20142d} or recovering the rotation matrices only~\cite{mahendran20173d}. 
Our expected goal is that by projecting the fine-grained 3D model according to the regressed pose estimation, the projection can align well with the object in the 2D image.
To tackle this problem, we integrate pose estimation into the Faster/Mask R-CNN framework~\cite{ren2015fasterrcnn,he2017mask} by sharing information between the detection and pose estimation branches.
However, a simple extension leads to inaccurate prediction result.
Therefore, we introduce dense 3D representation into the end-to-end deep framework named {\em 3D location field} that maps
each pixel to the 3D location on the model surface. The idea of using pixel-3D coordinates correspondences was explored on
multi-stage frameworks using RGB-D input \cite{taylor2012vitruvian,shotton2013coord,brachmann2014pose}. Under
end-to-end deep framework with RGB input at category-level, we show that this representation
can provide powerful supervision for the CNNs to efficiently capture the 3D shape of objects. Additionally, it requires no
rendering such that there is no domain gap between real-world annotated data and synthetic data.
Using large amount of synthetic location fields for pre-training, we overcome the problem of data shortage as well as
the domain gap caused by rendering.

Our contribution is three-fold. 
First, we collect a new large 3D pose dataset for fine-grained objects with a better match to the fine-detailed shapes of objects. 
Second, we propose a system based on Faster/Mask R-CNN that estimates a full perspective model parameters on our dataset. 
Third, we integrate {\em location field}, a dense 3D representation that efficiently encodes the object 3D shapes, into deep framework in an end-to-end fashion.
This goes beyond previous works on category-level pose estimation, which only estimate discrete/continuous viewpoint angles or recover rotation matrices often with the help of key points. 
\begin{table}
\begin{center}
\small
\begin{tabular}{|c|c|c|c|c|}
\hline
Dataset & \# class & \# image & annotation & fine-grained \\
\hline
3D Object \cite{savarese20073d} & 10 & 6,675 & discretized view & \xmark \\
EPFL Cars \cite{ozuysal2009pose} & 1 & 2,299 & continuous view & \xmark \\
Pascal 3D+ \cite{xiang2014beyond} & 12 & 30,899 & 2d-3d alignment & \xmark \\
ObjectNet3D \cite{xiang2016objectnet3d} & 100 & 90,127 & 2d-3d alignment & \xmark \\
\hline
StanfordCars 3D (Ours) & 196 & 16,185 & 2d-3d alignment & \cmark \\
CompCars 3D (Ours) & 113 & 5,696 & 2d-3d alignment & \cmark \\
Total (Ours) & 309 & 21881 & 2d-3d alignment & \cmark \\
\hline
\end{tabular}
\caption{\label{tb1} 
We provide a larger-scale pose annotation than most existing datasets. Although ObjectNet3D also annotates 100 classes with more than 90,000 images, their CAD models are for generic objects, not in fine-grained details.}
\end{center}
\end{table}

\section{Related Work}
\noindent\textbf{Dataset.} Earlier object pose datasets are limited not only in their dataset scales but also in the types of annotation they covered.
Table \ref{tb1} provides a quantitative comparison between our dataset and previous ones.
For example, 3D Object~\cite{savarese20073d} dataset only provides viewpoint annotation for 10 object classes with 10 instances for each class.
EPFL Car dataset~\cite{ozuysal2009pose} consists of 2,299 images of 20 car instances captured at multiple azimuth angles;
moreover, the other parameters including elevation and distance are kept almost the same for all the instances in order to simplify the problem~\cite{ozuysal2009pose}.
Pascal 3D+~\cite{xiang2014beyond} is perhaps the first large-scale 3D pose dataset for generic object categories, with 30,899 images from 12 different classes of the Pascal VOC dataset~\cite{everingham2010pascal}.
Recently, ObjectNet3D dataset~\cite{xiang2016objectnet3d} further extends the scale to 90,127 images of 100 categories.
Both Pascal 3D+ and ObjectNe3D datasets assume a camera model with 6 parameters to annotate.
However, different images in one hyper class (\ie, cars) are usually matched with a few coarse 3D CAD models, thereby the projection error might be large due to the lack of accurate CAD models in some cases.
Being aware of these problems, we therefore project fine-grained CAD models to match with images.
In addition, our datasets surpass most of previous ones in both scales of images and classes.

\noindent\textbf{Pose Estimation.} Despite the fact that continuous pose parameters are available for dataset such as Pascal 3D+, a majority of previous works \cite{xiang2014beyond,tulsiani2015viewpoints,pepik2012teaching,ghodrati20142d,su2015render} still casts the pose estimation problem as a multi-class classification of discrete viewpoint angles, which can be further refined as shown in \cite{yang2014object,hara2017designing}.
There are very few works except~\cite{pavlakos20176,mahendran20173d} that directly regresses the continuous pose parameters.
Although ~\cite{pavlakos20176} estimates a weak-perspective model for object categories and is able to lay the 3D models onto 2D images for visualization, its quantitative evaluation is still limited to 3D rotations. 
In contrast, we tackle a more challenging problem that estimates the full perspective matrices from a single image.
Our new dataset allows us to quantitatively evaluate the estimated perspective projection.
Based on this, we design a new efficient CNN framework as well as a new 3D representation that further improves the pose estimation accuracy.

\noindent\textbf{Fine-Grained Recognition.} Fine-grained recognition refers to the task of distinguishing sub-ordinate categories~\cite{wah2011cub,krause20133d,van2017inaturalist}. 
In earlier works, 3D information is a common source to gain recognition performance improvement~\cite{zia2013detailed,xiang2015data,mottaghi2015coarse,sochor2016boxcars}.
As deep learning prevails and fine-grained datasets become larger \cite{lin2015bilinear,krause2016unreasonable}, the effect of 3D information on recognition diminishes. 
Recently, \cite{sochor2016boxcars} incorporate 3D bounding box into deep framework when images of cars are taken from a fixed camera. 
On the other hand, almost all existing fine-grained datasets are lack of 3D pose labels or 3D shape information~\cite{krause20133d}, and pose estimation for fine-grained object categories are not well-studied. 
Our work fills this gap by annotating poses and matching CAD models on two existing popular fine-grained recognition datasets and performing the new task of pose estimation based on the augmented annotations.

\section{Dataset}
Our dataset annotation process is similar to ObjectNet3D~\cite{xiang2016objectnet3d}.
We first select the most appropriate 3D car model from ShapeNet~\cite{chang2015shapenet} for each category in the fine-grained image dataset.
For each image, we then obtain its pose parameters by asking the annotators to align the projection of the 3D model with the image using our designed interface.

\subsection{3D Models}
We build two fine-grained 3D pose datasets for vehicles. Each dataset consists of two parts, \ie, 2D images and 3D models. 
The 2D images of vehicles are collected from StanfordCars \cite{krause20133d} and CompCars \cite{yang2015large} respectively.
Target objects in most images are non-occluded and easy to identify.
In order to distinguish between fine-grained categories, we adopt a distinct model for each category.
Thanks to ShapeNet \cite{chang2015shapenet}, a large number of 3D models for fine-grained vehicles are available with make/model names in their meta data, which are used to find the corresponding 3D model given an image category name.
If there is no exact match between a category name and meta data, we manually select a visually similar 3D model for that category. 
For StanfordCars, we annotate images for all 196 categories, where 148 categories have exact matched models. 
For CompCars, we only include 113 categories with matched 3D models in ShapeNet.
To our best knowledge, our dataset is the very first one which employs fine-grained category aware 3D model in 3D pose estimation.

\subsection{Camera Model}
The world coordinate system is defined in accordance with the 3D model coordinate system. 
In this case, a point $\bf
X$ on a 3D model is projected onto a point $\bf x$ on a 2D image:
\begin{equation}\label{eq1}
{\bf x} = {\cal{P}}{\bf X},
\end{equation}
via a perspective projection matrix:
\begin{equation}\label{eq2}
{\cal P} = K\left[ {R|T} \right],
\end{equation}
where $K$ denotes the intrinsic parameter:
\begin{equation}\label{eq3}
K = \left[ {\begin{array}{*{20}{c}}
f&0&{{u}}\\
0&f&{{v}}\\
0&0&1
\end{array}} \right],
\end{equation}
and $R$ encodes a $3 \times 3$ rotation matrix between the world and camera coordinate systems, parameterized by three angles, i.e., elevation $e$, azimuth $a$ and in-plane rotation $\theta$.
We assume that the camera is always facing towards the origin of the 3D model. Hence the translation $T = [0, 0, d]^{\rm T}$ is
only defined up to the model depth $d$, the distance between the origins of two coordinate systems, and the principal point
$(u, v)$ is the projection of the origin of world coordinate system on the image. As a result, our model has 7
parameters in total: camera focal length $f$, principal point location $u$, $v$, azimuth $a$, elevation $e$,
in-plane rotation $\theta$ and model depth $d$. Note that, since
the images are collected online, even the annotated intrinsic parameters ($u$, $v$ and $f$) are approximation. Compared with previous annotations \cite{xiang2014beyond,xiang2016objectnet3d} with 6 parameters ($f$ fixed), our camera model considers both the camera focal length $f$ and object depth $d$ in a full perspective projection for finer 2D-3D alignment.
\begin{figure}
\centering
\includegraphics[width=\textwidth]{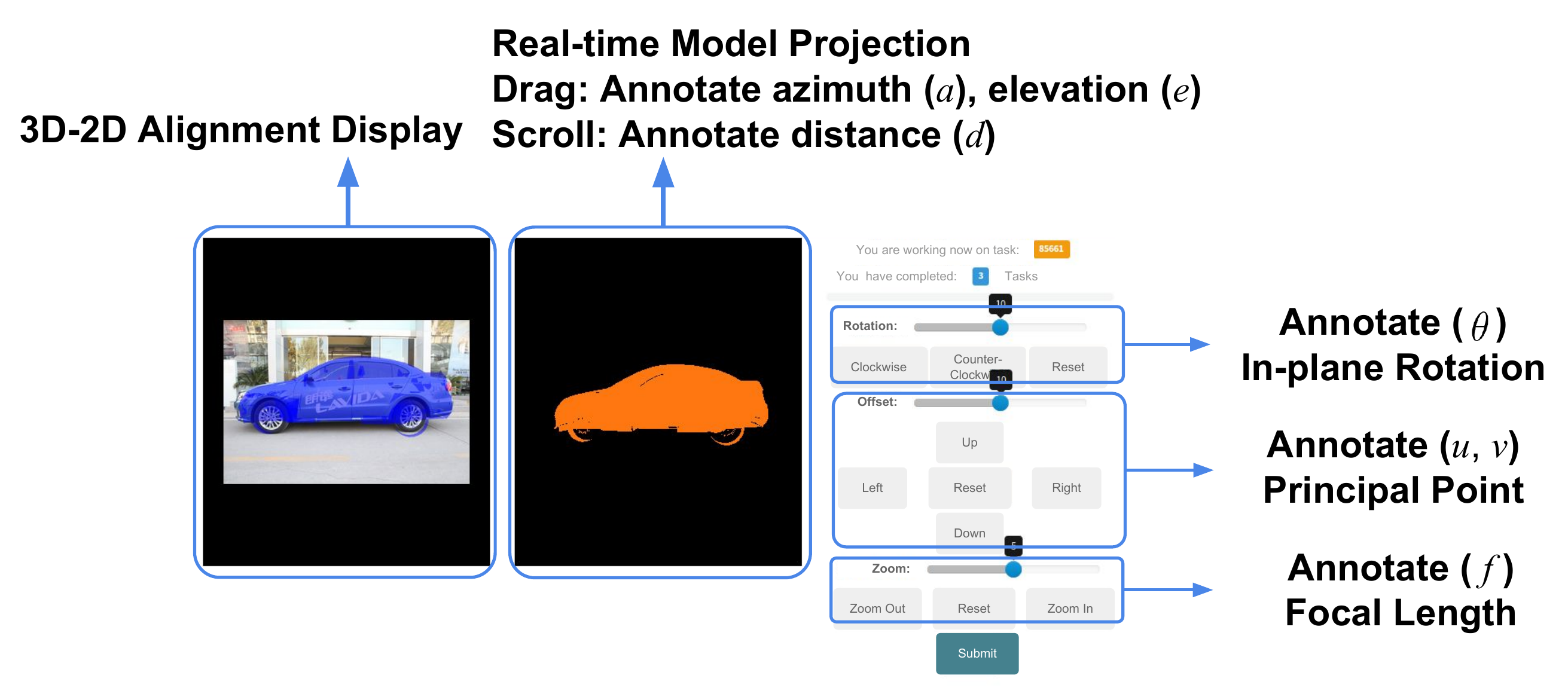}
\caption{\label{fig8}An overview of our annotation interface. 
}
\end{figure}

\subsection{2D-3D Alignment}
We annotate 3D pose information for all 2D images in our datasets through crowd-sourcing. 
To facilitate the annotation process, we develop an annotation tool illustrated in Figure \ref{fig8}. 
For each image during annotation, we choose the 3D model according to the fine-grained car type given beforehand. Then, we ask the annotators to adjust the 7 parameters so that the projected 3D model is aligned with the target object in 2D image. 
This process can be roughly summarized as follows: (1) shift the 3D model such that the center of the model (the origin of the world coordinate system) is roughly aligned with the center of the target object in the 2D image; (2) rotate the
model to the same orientation as the target object in the 2D image; (3) adjust the model depth $d$ and camera focal length $f$ to match the size of the target object in the 2D image. Some finer adjustment might be applied after the three main steps. In this way we annotate all 7 parameters across the whole dataset. On average, each image takes approximately 60 seconds to annotate by an experienced annotator. To ensure the quality, after one round of annotation across the whole dataset, we perform quality check and let the annotators do a second round revision for unqualified examples.
\begin{figure}
\centering
\begin{tabular}{ccc}
\includegraphics[width=0.3\textwidth]{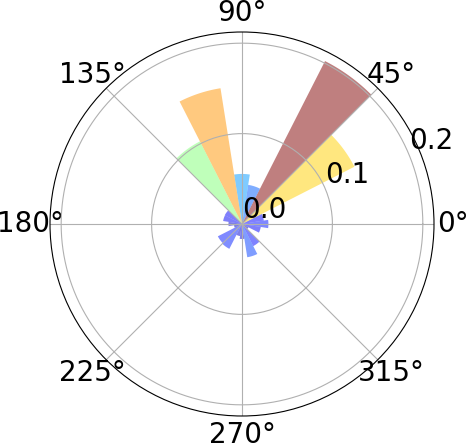} &
\includegraphics[width=0.3\textwidth]{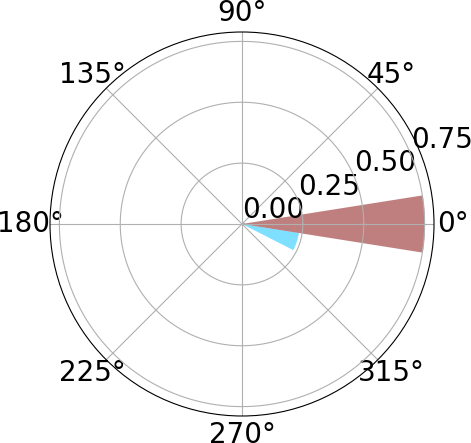}
&
\includegraphics[width=0.3\textwidth]{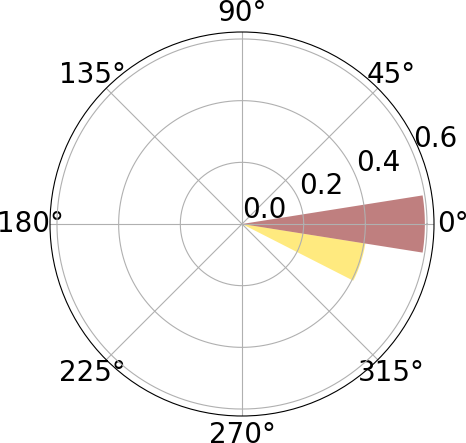} \\
azimuth & elevation & theta
\end{tabular}
\caption{\label{fig99}The polar histogram of the three key pose parameters in our annotated StanfordCars 3D dataset.}
\end{figure}

\begin{figure}
\centering
\begin{tabular}{ccc}
\includegraphics[width=0.3\textwidth]{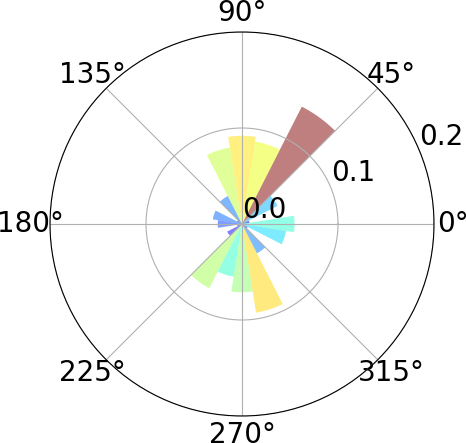} &
\includegraphics[width=0.3\textwidth]{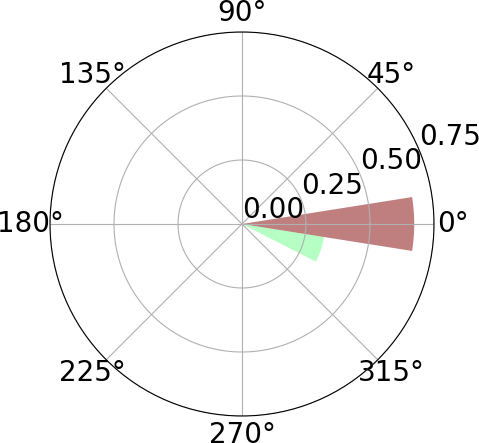}
&
\includegraphics[width=0.3\textwidth]{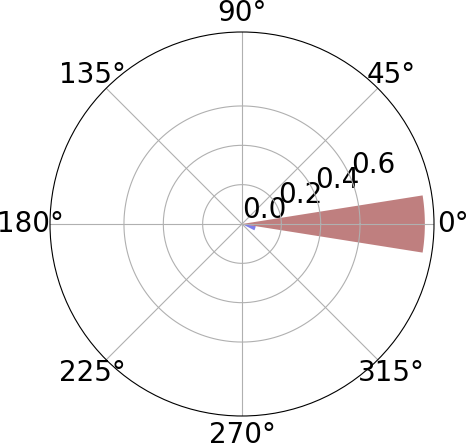} \\
azimuth & elevation & theta
\end{tabular}
\caption{\label{fig98}The polar histogram of the three key pose parameters in our annotated CompCars 3D dataset.}
\end{figure}

\subsection{Dataset Statistics}
We plot the distributions of azimuth ($a$), elevation ($e$) and in-plane rotation ($\theta$) in Figure \ref{fig99} and Figure \ref{fig98} for StanfordCars 3D and CompCars 3D, respectively. For azimuth, due to the nature of the original fine-grained recognition dataset, we found it is not uniformly distributed, while the distributions of the two dataset are complementary to some degree. Elevations and in-plane rotations are not severe as expected, since the images of cars are usually taken from the ground view.

\section{3D Pose Estimation for Fine-Grained Object Categories}\label{sec4}
Given an input image of a fine-grained object, our task is to predict \textit{all} the 7 parameters related to Equation (\ref{eq2}), \ie, 3D rotation $R(a, e, \theta)$, distance $d$, principal point $(u, v)$ and $f$, such that the projected 3D model can align well with the object in the 2D image.

\subsection{Baseline Framework}\label{sec4_1}
Our baseline method only uses 2D appearance to regress the pose parameters.
It is a modified version of Faster R-CNN \cite{ren2015fasterrcnn} which was originally designed for object detection. 
Casting our pose estimation problem into a detection framework is motivated by the relation between the two tasks.
Since we are not using key points as an attention mechanism, performing pose
estimation within the region of interest (RoI) helps us get rid of unrelated image regions hence make use of 2D information more effectively. 
In addition, 3D pose estimation is highly related to the detection task, especially the intrinsic parameters in Equation (\ref{eq3}).

We parametrize the 3D rotation using the \textit{quaternion} representation, converted from the angles $(a, e, \theta)$. The principal point $(u, v)$ is highly related to RoI center. Therefore, we regress $(\Delta u, \Delta v)$,
the offset of the principal point from the RoI center. Such offset exists since the projection of the 3D object center
might not necessarily be the 2D center depending on the poses. For other parameters ($d$ and $f$), we regress the standard format as they are.

\begin{figure}
\centering
\includegraphics[width=\textwidth]{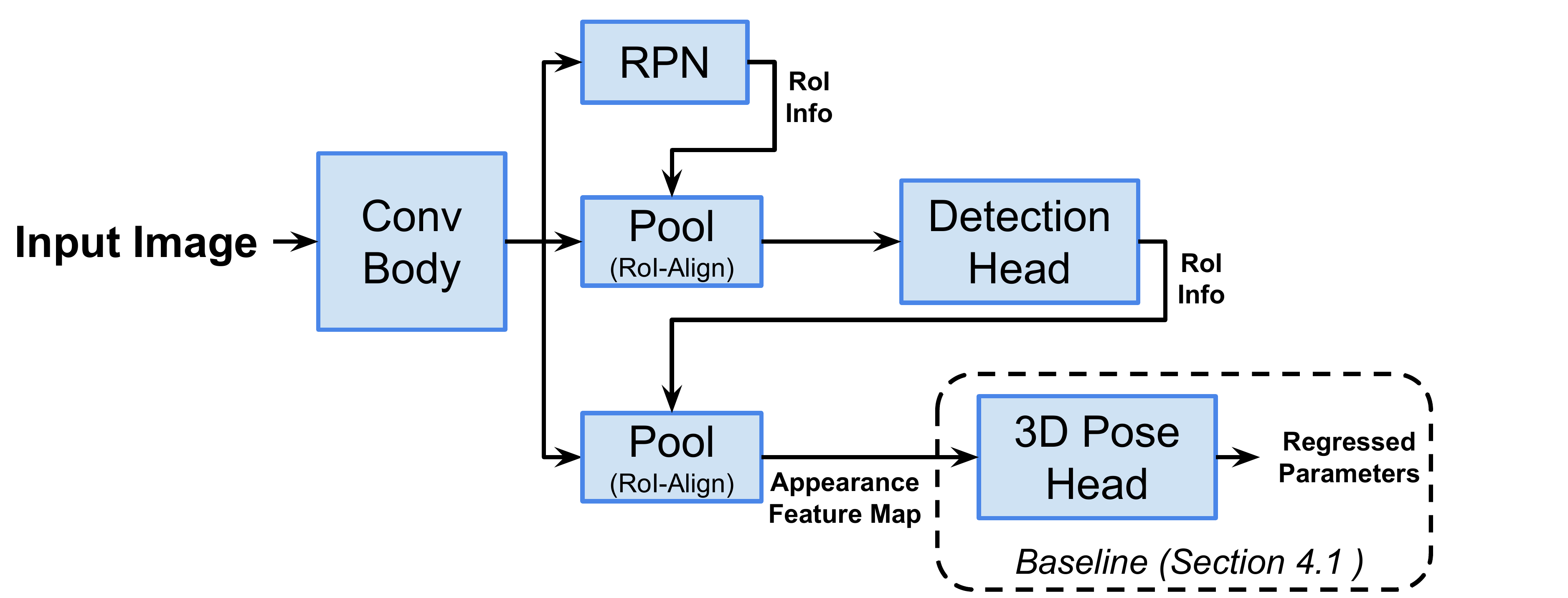}
\caption{\label{fig2}Our base pose estimation framework. Given an input 2D image, we adapt the Mask R-CNN framework to regress the pose parameters from the pooled appearance feature map with its area determined by the Detection module. The whole network is trained end-to-end.}
\end{figure}

The modification of the network architecture is relatively straightforward. As shown in Figure \ref{fig2}, we add a pose estimation branch along with the existing class prediction and bounding box regression branches. Similar to the bounding box regression branch, the estimation of each group of pose parameters consists of a fully-connected (FC) layer and a smoothed $L_1$ loss. The centers of the RoIs are also used to adjust the regression targets at training time and generate the final predictions at test time, as discussed above. For each training image, its bounding box is figured out from the
perspective projection of the corresponding 3D model. Since we have fine-grained 3D models and high-quality annotations, these bounding boxes are tight to their corresponding objects.

\subsection{Improve Pose Estimation via 3D Location Field}\label{sec4_2}
The key difference of our dataset to previous ones is that we have fine-grained 3D models such that the projection aligns better with the image. 
This advantage allows us to explore the usage of dense 3D representations in addition to 2D appearance to regress the pose parameters.

\begin{figure}
\centering
\includegraphics[width=\textwidth]{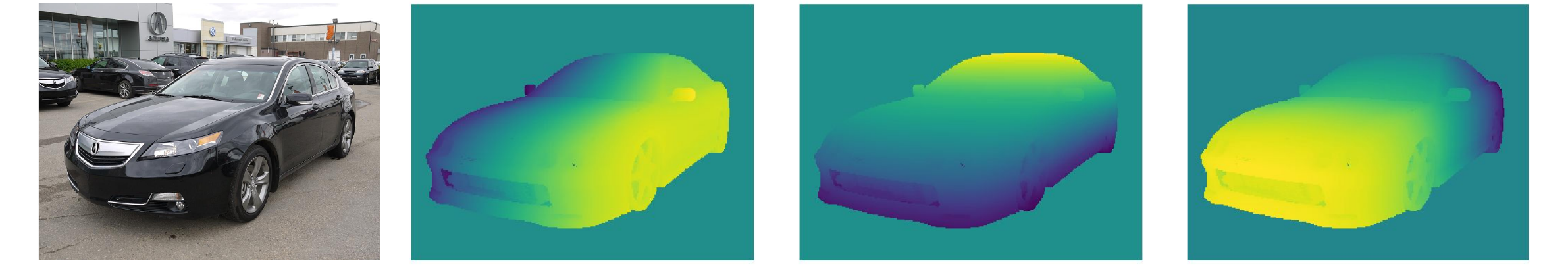}
\caption{\label{fig3}A sample image and its corresponding 3D location field. The location field is a 3-dimensional tensor with the same size as an image. The last channel encodes the 3d locations of a pixel on the visible surface of the 3D model.}
\end{figure}

Given an object in an image and its 3D model, our representation, named as {\em 3D location field}, maps every foreground pixel to its corresponding
location on the surface of the 3D model, \ie, $f(x, y) = (X, Y, Z)$. The resulting field has the same size as the image and has three channels containing the $X$, $Y$ and $Z$ coordinates respectively. A sample image with corresponding 3D
location field can be seen in Figure \ref{fig3}. The 3D location field is a dense
representation of 3D information which can be directly used as network input.

\begin{figure}
\centering
\includegraphics[width=\textwidth]{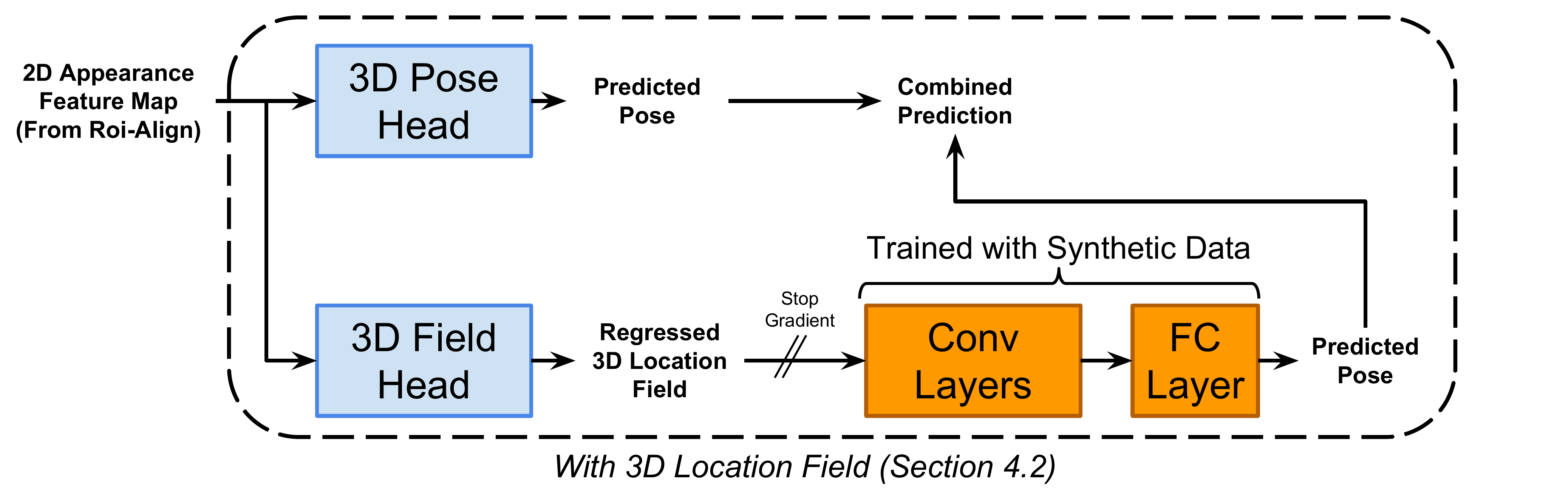}
\caption{\label{fig4}Our improved network architecture of using 3D location field to help pose estimation. The block in the dash-line box is to replace the corresponding base network in Figure \ref{fig2}. The key difference is that we add a 3D Field branch that also estimates the pose parameters.}
\end{figure}

We explore the usage of 3D location field to improve pose estimation based on Mask R-CNN.
We would still expect only 2D image input at test time, therefore we regress 3D location field and use the regressed field for pose estimation. 
Based on the framework in Figure \ref{fig2}, we add a branch to regress 3D location field (instead of regressing binary masks in Mask R-CNN). 
The regressed location fields are fed into a CNN consisting of additional convolutional layers followed by layers to regress the pose parameters. 
The regressions from 2D appearance (as part of Figure \ref{fig2}) and 3D location field are later combined to produce the final pose parameters. 
Figure \ref{fig4} shows the detailed network structure.

We train the pose regression from location fields using a large amount of synthetic data. 
The synthetic location fields are generated from the 3D models with various pre-defined poses. 
The location field is a very suitable representation for synthetic data augmentation due to the following reasons: 
(\textit{i}) the field only encodes 3D location information without any rendering of 3D models and naturally avoids the domain gap between synthetic data and
photo-realistic data; 
(\textit{ii}) the field is invariant to color, texture and scale of the images.

\section{Experiments}\label{sec5}

\subsection{Evaluation Metrics}\label{sec5_1}
For each test sample, we introduce two metrics to comprehensively evaluate object poses.

Following \cite{tulsiani2015viewpoints,pavlakos20176}, the first metric, \textbf{Rotation Error}, focuses on the quality of viewpoint estimation only. Given
the predicted and ground truth rotation matrices $\{R, R_{\rm gt}\}$, the difference between the two measured by
geodesic distance is $e_{\rm R} = \frac{1}{\sqrt{2}} \|\log(R^{\rm T}R_{\rm gt})\|_{\rm F}$.

The second metric evaluates the overall quality of perspective projection. Our evaluation
metric is based on \textbf{Average Distance of Model Points} in \cite{hodan2016eval3d}, which measures the averaged
distance between predicted projected points and their corresponding ground truth projections. Concretely, given one test
result $\{\cal{P}, \cal{P}_{\rm
gt}, \cal{M}\}$, where its predicted pose is $\cal{P}$, its ground truth pose $\cal{P}_{\rm gt}$ and corresponding 3D model
$\cal{M}$, the metric is defined as
\begin{align}
e_{\rm ADD}(\cal{\cal{P}, \cal{P}_{\rm gt}; \cal{M}}) = \underset{\bf{X}\in\cal{M}} {\rm avg} \left\|\cal{P} \bf{X} -
\cal{P}_{\rm gt} \bf{X}\right\|_{\rm 2}
\end{align}
According to \cite{hodan2016eval3d}, this is the most widely-used error function to evaluate a projection matrix. 
The unit of the above distance is the number of pixels. To make the metric scale-invariant, we normalize it using the
diameter of the 2D bounding box. We denote the normalized distance as $\tilde{e}_{\rm ADD}$. It is worth mentioning again that
the 3D models are only used when computing the evaluation metrics. During test time, only a single 2D image is fed into
the network to predict the pose $\cal{P}$.

To measure the performance over the whole test set, we compute the mean and median of $e_R$ and $\tilde{e}_{\rm ADD}$ over all test
samples. Also, by setting thresholds on the two metrics, we can get an accuracy number. For $e_R$, following \cite{tulsiani2015viewpoints,pavlakos20176}, we set the threshold to be $\frac{\pi}{6}$. For $\tilde{e}_{\rm ADD}$, the common threshold is 0.1,
which means that the prediction with average projection error less than 10\% of the 2D diameter is considered correct.

\subsection{Experimental Settings}\label{sec5_2}
\noindent\textbf{Data Split.} For StanfordCars 3D, since we have annotated all the images, we follow the standard train/test split provided by the original dataset \cite{krause20133d} with 8144 training examples and 8041 testing examples. For CompCars 3D, we randomly sample $2/3$ of our annotated data as training set and the rest $1/3$ as testing set, resulting in 3798 training and 1898 test examples.

\noindent\textbf{Baseline Implementation.} Our implementation is based on the Detectron package \cite{detectron},
which includes Faster/Mask R-CNN implementations. The convolutional body (\ie, the
``backbone'' in \cite{he2017mask}) used for the baseline is ResNet-50. For fair comparison, the convolutional body is
initialized from ImageNet pre-trained
model, and other layers are randomly initialized (\ie, we are not using COCO pre-trained detectors). 
Following the setting of Mask R-CNN, the whole network is trained end-to-end. 
At test time, we adopt a cascaded strategy, where the 3D pose branch is applied only to the highest scoring box
prediction. 

\noindent\textbf{Comparison to Previous Baselines.} It is worth mentioning that, when only evaluating the rotation error in
Section \ref{sec5_1}, our baseline in Figure \ref{fig2} is almost identical to the baselines in Pascal3D+ \cite{xiang2014beyond} and ObjectNet3D
\cite{xiang2016objectnet3d} except that their detection and pose estimation heads are parallel while ours is cascaded.

\noindent\textbf{3D Location Field.} In Section
\ref{sec4_2}, incorporating 3D location fields involves two steps -- field regression and pose regression from fields.
Field regression is trained together with detection and baseline pose estimation in an end-to-end fashion, similar to
Mask R-CNN. The ground truth training fields are generated from the annotations (3D models and poses). The second step,
pose regression from fields is trained using the synthetic data generated from the pool of matched 3D models in a
dataset (38102/14017 synthetic samples for StanfordCars\&CompCars 3D). We only regress the quaternion using the location fields.
\begin{table}
\small
\centering
\begin{tabular}{|c|c|c|c||c|c|c|}
\hline
Method & Median $e_{\rm R}$ & Mean $e_{\rm R}$& $Acc_{\rm \frac{\pi} {6}}$ & Median $\tilde{e}_{\rm ADD}$ & Mean $\tilde{e}_{\rm ADD}$ & $Acc_{\rm th=0.1}$ \\
\hline
Baseline& 6.68 & 9.89 & 96.59 & 0.0888 & 0.1087 & 60.04\\
w./ Field & \bf{5.68} & \bf{7.67} & \bf{98.73} & \bf{0.0834} & \bf{0.0977} & \bf{66.07}\\
\hline
\end{tabular}
\caption{\label{tb3}Experimental results on StanfordCars 3D dataset. The two rows show the baseline results (Section \ref{sec4_1}) and the results with 3D location field (Section \ref{sec4_2}), respectively. The rotation error $e_{\rm R}$ is measured in
degree ($^{\circ}$). The accuracy ($Acc_{\rm \frac{\pi} {6}}$ and $Acc_{\rm th=0.1}$) is measured in percentage (\%). Please see Section \ref{sec5_1} for details about evaluation metrics.}
\end{table}
\begin{table}
\small
\centering
\begin{tabular}{|c|c|c|c||c|c|c|}
\hline
Method & Median $e_{\rm R}$ & Mean $e_{\rm R}$& $Acc_{\rm \frac{\pi} {6}}$ & Median $\tilde{e}_{\rm ADD}$ & Mean $\tilde{e}_{\rm ADD}$ & $Acc_{\rm th=0.1}$ \\
\hline
Baseline& 8.09 & 13.02 & 93.62 & 0.1275 & 0.1580 & 32.52\\
w./ Field & 6.14 & 8.98 & 98.00 & 0.1141 & 0.1408 & 40.15\\
\hline
FT Baseline& 5.51 & 8.69 & 96.84 & 0.0878 & 0.1123 & 58.58\\
FT w./ Field & \bf{4.74} & \bf{7.45} & \bf{98.31} & \bf{0.0836} & \bf{0.1047} & \bf{64.01}\\
\hline
\end{tabular}
\caption{\label{tb4}Experimental results on CompCars 3D dataset. The last two rows show results fine-tuned (FT) from a StanfordCars 3D pre-trained model.}
\end{table}

\subsection{Results and Analysis}\label{sec5_3}
\begin{figure}
\centering
\includegraphics[width=0.45\linewidth]{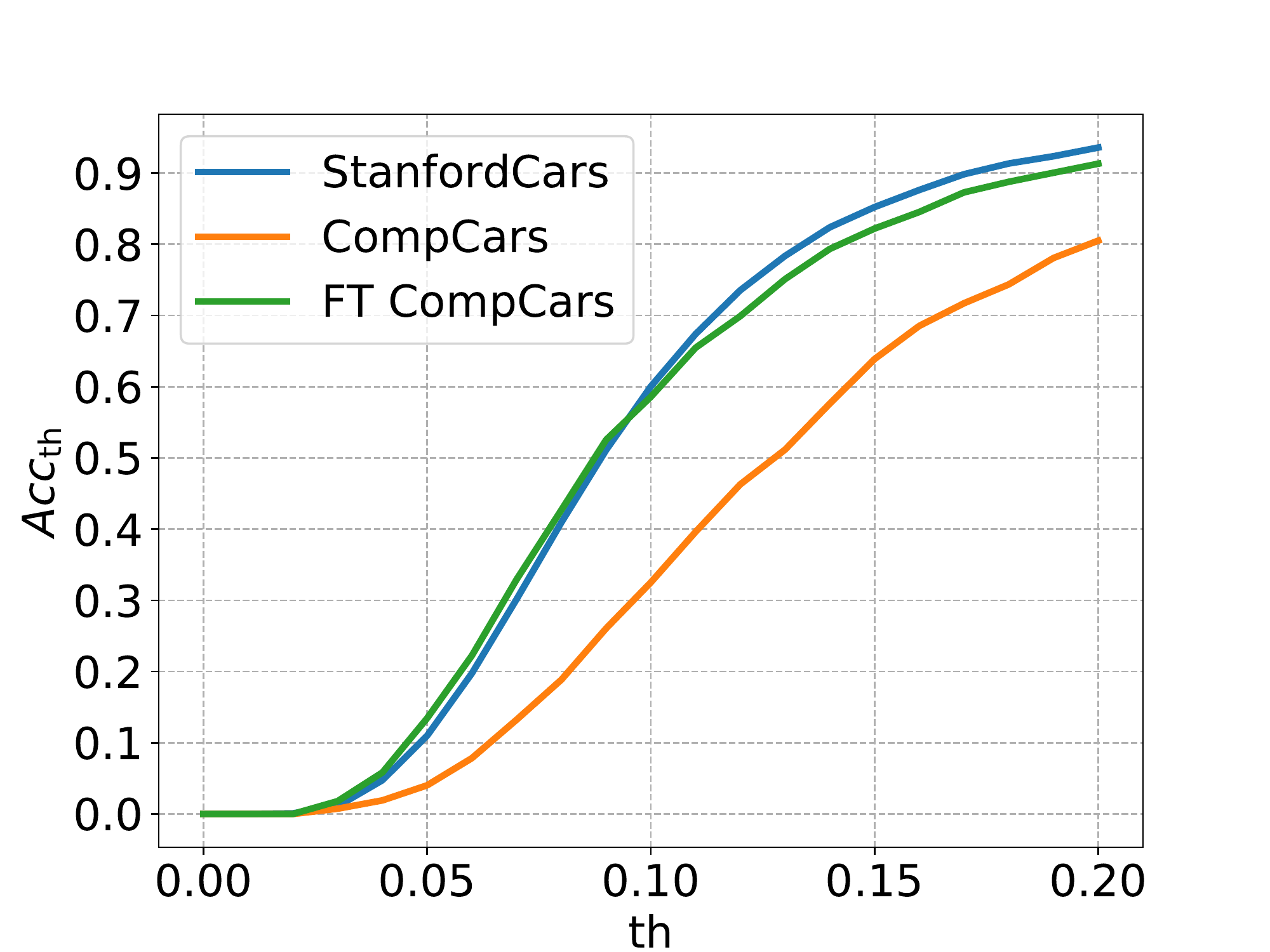}
\includegraphics[width=0.45\linewidth]{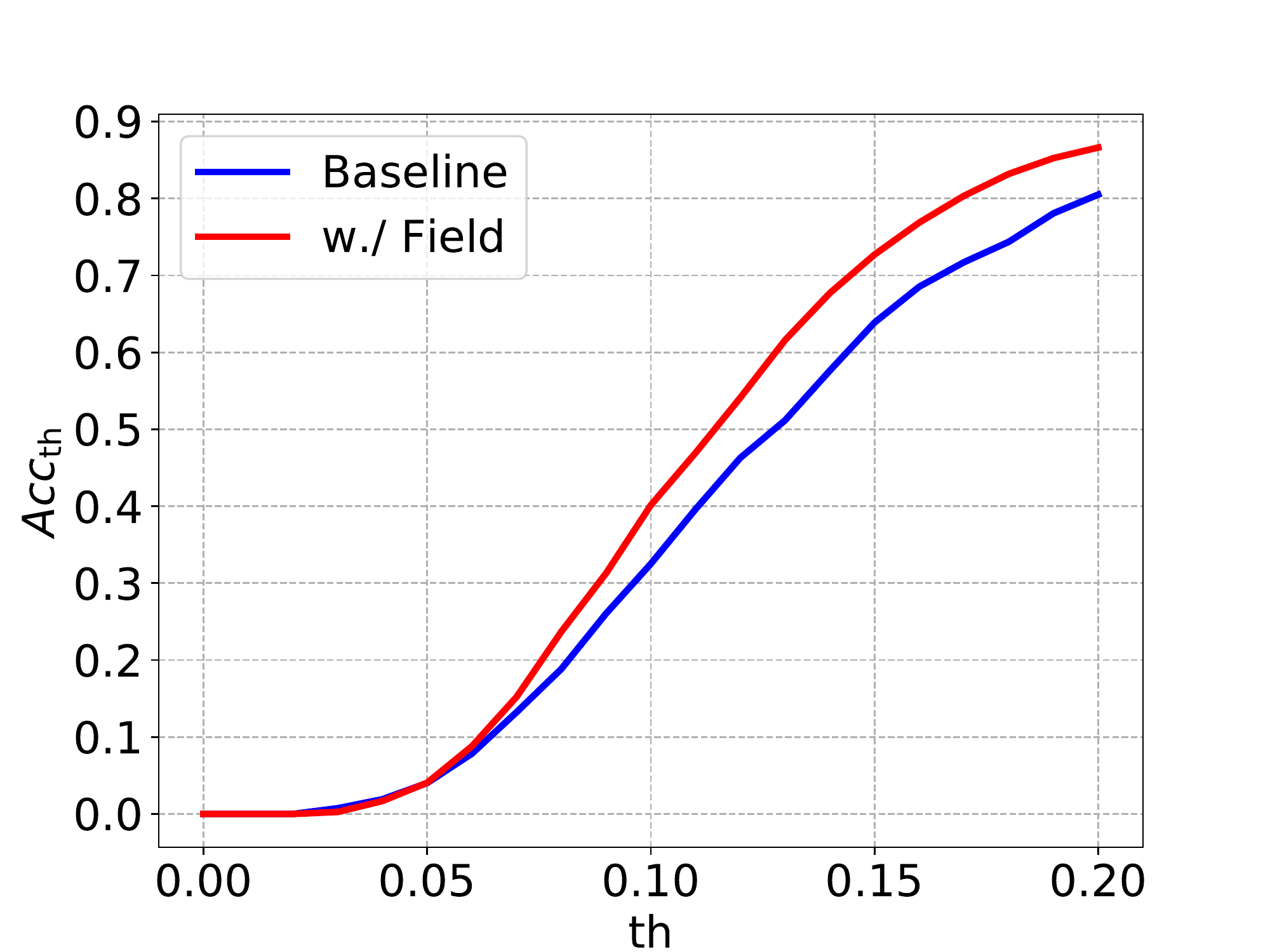}
\caption{\label{fig6}Left: Plot of $Acc_{\rm th}$ \textit{w.r.t.} threshold for the three baselines. Right: For CompCars 3D, compare the result using location field to the baseline curve.}
\end{figure}

The quantitative results for StanfordCars 3D and CompCars 3D are shown in Table
\ref{tb3} and Table \ref{tb4} respectively. The changes of $Acc_{\rm th}$ \textit{w.r.t} the threshold for the
datasets are shown in Figure \ref{fig6}. For CompCars 3D dataset, besides ImageNet
initialization we also report the result finetuned from a StanfordCars 3D pretrained model, since the number of training samples in StanfordCars is relatively larger. 

As can be seen in Table \ref{tb3} and \ref{tb4}, our baseline performs very well on estimating the rotation matrix for
both datasets, with Median $e_R$ less than 10 degrees and
$Acc_{\frac{\pi}{6}}$ around 95\%. While recovering the full perspective model is a much more challenging
task, Table \ref{tb3} shows that promising performance can be achieved with enough properly annotated training
samples. For StanfordCars 3D, Median $\tilde{e}_{\rm ADD}$ (the median of the
average projection error) is less than 10\% of the diameter of the 2D bounding box. When the training set is limited, from the first and the third row of Table \ref{tb4}, we can see the effectiveness of transfer learning from a larger
dataset. Regarding the effectiveness of the 3D location field, we can observe consistent performance gain across all
datasets. The main reasons are two-fold: (\textit{i}) this 3D representation enables the usage of large amounts of synthetic training data with no domain gap; (\textit{ii}) our field regression adapted from Mask R-CNN works well such that even the pose prediction based on the regressed field can help a lot at test time.
\begin{figure}
\includegraphics[width=1.02\linewidth]{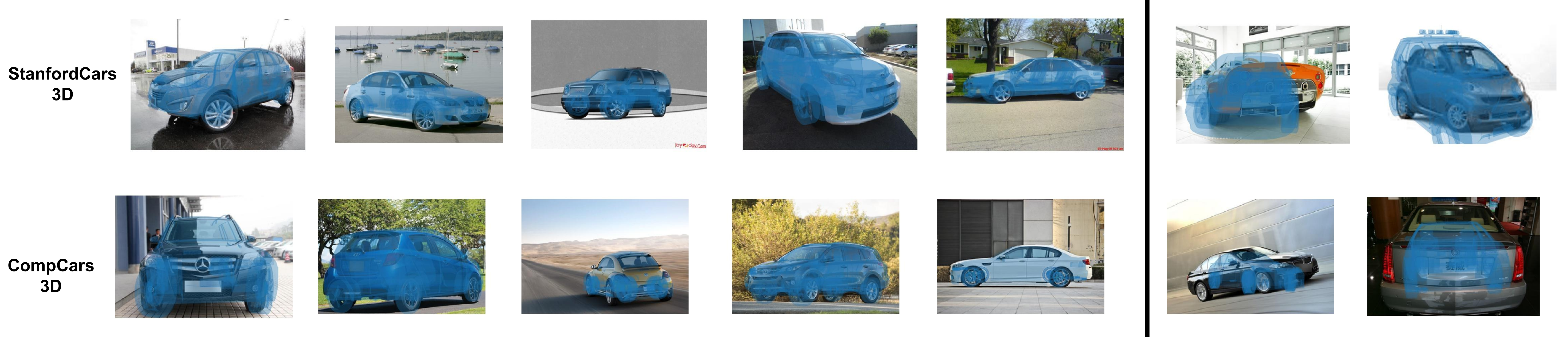}
\caption{\label{fig5}Visualizations of predicted poses for test examples. For each dataset, we show five examples of successful predictions and two of the failure cases, separated by the solid black line in the figure.}
\end{figure}

We visualize the predicted poses in Figure \ref{fig5}.
As shown on the left part of Figure \ref{fig5}, our method is able to handle poses of various orientations, scales and locations of the projection. 
On the right part of Figure \ref{fig5}, failure cases exist in our predictions, indicating there are still potential rooms for improvement, especially for the estimation of scale, cases with large perspective distortion and some uncommon poses with few training samples.

\section{Conclusion}\label{sec7}
We study the problem of pose estimation for fine-grained object categories. We annotate two popular fine-grained recognition datasets with fine-grained 3D shapes and poses.
We propose an approach to estimate the full perspective parameters from a single image.
We further propose 3D location field as a dense 3D representation to facilitate pose estimation.
Experiments on our datasets suggest that this is an interesting problem in future.

\bibliographystyle{splncs04}
\bibliography{egbib}
\end{document}